%% file: arxiv.tex
\documentclass[twoside]{article}

\usepackage{arxiv}

\usepackage{amsmath,amsthm,verbatim,amssymb,amsfonts,amscd, graphicx}
\usepackage{graphics}
\usepackage{centernot}

\theoremstyle{plain}

\theoremstyle{definition}

\usepackage{wrapfig}

\usepackage{caption}
\usepackage{subcaption}
\usepackage{helvet}  %
\usepackage{courier}  %
\usepackage{url}  %
\usepackage{graphicx}  %
\usepackage{multirow}
\usepackage{amsthm}
\usepackage{color}
\usepackage{MnSymbol}
\usepackage{makecell}
\usepackage{arydshln}
\usepackage{amsmath}
\usepackage{xcolor}
\usepackage{caption} 
\usepackage[numbers]{natbib}
\usepackage{textcomp}
\usepackage{wrapfig}
\usepackage{algorithm}
\usepackage{algorithmic}

\begin{document}

\title{An Investigation of how Label Smoothing Affects Generalization}

\author{Blair Chen$^*$ \And Liu Ziyin$^*$ \And Zihao Wang \And Paul Pu Liang}
\author{Blair Chen\thanks{Equal contribution.} \\
	Carnegie Mellon University\\
	\And
	Liu Ziyin$^*$\\
	University of Tokyo\\
	\And
	Zihao Wang\\
	Hong Kong University\\ of Science and Technology\\
	\And
	Paul Pu Liang\\
	Carnegie Mellon University\\
}

\maketitle

\begin{abstract}
It has been hypothesized that label smoothing can reduce overfitting and improve generalization, and current empirical evidence seems to corroborate these effects. However, there is a lack of mathematical understanding of when and why such empirical improvements occur. In this paper, as a step towards understanding why label smoothing is effective, we propose a theoretical framework to show how label smoothing provides in controlling the generalization loss. In particular, we show that this benefit can be precisely formulated and identified in the label noise setting, where the training is partially mislabeled.
Our theory also predicts the existence of an \textit{optimal label smoothing} point, a single value for the label smoothing hyperparameter that minimizes generalization loss. Extensive experiments are done to confirm the predictions of our theory. We believe that our findings will help both theoreticians and practitioners understand label smoothing, and better apply them to real-world datasets.
\end{abstract}

\section{Introduction}
\label{sec: label smoothing definition}

Label smoothing has emerged as a useful technique in training neural networks~\citep{goodfellow2016deep,szegedy2016rethinking,shafahi2019label,lukasik2020does,zoph2018learning, real2019regularized} and empirical results have pointed towards its ability to improve performance across a suite of tasks in image classification, machine translation, and language modeling~\citep{muller2019does} (see Table~\ref{tab: survey} for a summary of positive empirical results). Given a classification problem where the training targets $y_i$ take the form of a one-hot vector ($1$ for the target class index and $0$ elsewhere), label smoothing uses a smoothing hyperparameter $p$ to define a ``soft'' version of these training targets:
\begin{align}
    y_i &:= \underbrace{(0,...,0, 1,0,...,0)}_{M \text{ classes}} \to \hat{y}_i := \left( \frac{1-p}{M-1},..., \frac{1-p}{M-1}, p, \frac{1-p}{M-1},..., \frac{1-p}{M-1} \right).
\end{align}
However, due to the non-linear nature of neural networks, it is difficult to understand exactly how label smoothing works in improving training. Recent work has suggested that it helps to learn a more separable representation space~\citep{muller2019does}, penalizes over-confident inputs~\citep{pereyra2017regularizing}, and is particularly useful when label noise is present~\citep{lukasik2020does}, while cautioning that it is unfavorable to sparse distributions~\citep{meister2020generalized} and decreases robustness to adversarial attacks~\citep{zantedeschi2017efficient}. While these findings have shed some empirical insight on the benefits and drawbacks of applying label smoothing, there has not been any theoretical guidance as to how the label smoothing parameter $p$ should be chosen. Intuitively, if label smoothing functions like a regularization term, then a tradeoff between expressivity of the model and regularization should exist and an optimal value of $p$ is expected, which we will justify in this work.

The lack of guidance in choosing the hyperparameter $p$ is caused by not having a theoretical framework to formalize the benefit of label smoothing. We show that this benefit can be precisely formulated and identified in the label noise setting, where the training is partially mislabeled and thus, might take a different distribution from that of the test loss \citep{natarajan2013learning}. %
\input{tables/literature_survey}

Our theoretical results are based on interpreting label smoothing as a regularization technique and quantifying the tradeoffs between estimation and regularization. These results also allow us to predict where the \textit{optimal label smoothing} point lies for the best performance.
We test these theoretical predictions by performing extensive experiments on different models, random seeds, and datasets, showing that label smoothing can indeed be used to control generalization loss.

\section{Notation and Problem Setting}
\label{loss_s1}

This section sets up the notation and presents the background for our theoretical analysis. We begin by providing more background on related interpretations of label smoothing analyzed in recent literature before analyzing the generalization error of a model trained with label smoothing.

\textbf{Background:} Label-smoothing is a regularization technique that is only applicable to classification problems; more specifically, it only applies to the negative log-likelihood loss (nll) and its closed related variants that takes a form similar to KL divergence, such as the gambler's loss \citep{liu2019deep}.
Label smoothing can be seen as a element-wise function on one-hot targets $y_i$ that maps $0$ to $\frac{1-p}{M-1}$ and $1$ to $p$. This function can be equivalently represented by a label smoothing matrix $S_p$, whose diagonal elements are $p$ and the off-diagonal elements are $\frac{1-p}{M-1}$. %
We say that such label smoothing scheme performs \textit{uniform smoothing} (since the off-diagonal elements are the same). It is conceivable that the off-diagonal elements are non-uniform; one can define, for a general probability transition matrix $S$, $S-$label smoothing as
\begin{equation}
    \hat{y}_i = Sy_i,
\end{equation}
where $y_i$ is the original target (be it one-hot or not), and $\hat{y}_i$ is the smoothed target. While existing literature provides no guidance about where and when such a generalized label smoothing matrix can be helpful, we provide a theory for the advantages of making this generalization in Section~\ref{sec: non-uniform transition matrix}. Also, while the hyperparameter $p$ is crucial for the success of label smoothing, there is no work yet that gives any guidance for choosing $p$. Our theoretical results will shed further practical insights for choosing $p$.

\textbf{Analysis of the training objective:} %
Consider a learning task with $N$ input-targets pairs $\{(x_i, y_i)\}_{i=1,...,N}$ forming the test set. %
We assume that $(x_i,\ y_i)$ are drawn i.i.d. from a joint distribution $p(x,y)= p(y|x)p(x)$. We also assume that, during training, for any given $x_i$, $y_i\in \{0, 1\}$ can be uniquely determined, so that $p(y|x) \in \{0, 1\}$. We also assume that the distribution of two classes is balanced, i.e. $p(y=1) = p(y=0) = 1/2$. We denote model outputs as $f(x_i):=f_i$. The \textit{empirical} generalization loss $\ell_N[f]$ is defined as
{\fontsize{9.5}{12}\selectfont
\begin{align*}
    \ell_N[f] &=  -\sum_{y_i = 1} p(x_i, y_i) \log(f_i)-  \sum_{y_i = 0} p(x_i, y_i) \log (1 - f_i)\\
    &= -\frac{1}{N} \sum_{y_i = 1} y_i \log (f_i) - \frac{1}{N} \sum_{y_i = 0} (1 - y_i) \log (1 - f_i)
\end{align*}
}which is the cross-entropy loss on the test set. As in previous works, we focus on studying the minimizer of the training error and the generalization ability of such a minimizer~\citep{arora2019theoretical}. This approach characterizes neural networks well since they are frequently overparametrized and are shown to have the ability to achieve zero training loss given sufficient training time~\citep{Zhang_rethink, du2018gradient}. %
A $p-$label smoothing is then defined as setting $p(y_i|x=i) = (p, 1-p)$. The optimal solution in this case is simply $f_i = \mathbb{E} [y_i] = p$, where the generalization loss converges to
\begin{equation}
    \ell^* = -p\log p - (1-p)\log(1-p) = H(p);
\end{equation}
as $p\to 1$, the generalization error converges to $0$.

Now, we assume that label noise is present in the dataset such that each label is flipped to the other label with probability $r:= 1-a$, where $r<0.5$. We define $r$ to be the \textit{corruption rate} and $a$ to be the \textit{clean rate}. The generalization error of the new training loss minimizer becomes
{\fontsize{9.5}{12}\selectfont
\begin{align}
\label{eq: training loss}
    \Tilde{\ell}[f] %
    &= -\frac{1}{N}\left[ \sum_{y_i=1, \Tilde{y}_i = 1} \log (f_i)  + \sum_{y_i=1, \Tilde{y}_i = 0}  \log (1 - f_i)\right] -\frac{1}{N}\left[ \sum_{y_i=0, \Tilde{y}_i = 1} \log (f_i) + \sum_{y_i=0, \Tilde{y}_i = 0} \log (1 - f_i)\right]\\\\
    &= \ell_1[f] + \ell_0[f]\label{eq: training loss 2}
\end{align}
}where $\Tilde{y}$ denotes the new set of perturbed labels. Therefore, we have partitioned the original loss function into two separate loss functions, where $\ell_0$ is the loss for the data points whose original label is $0$, and likewise for $\ell_1$. %

\begin{figure*}
    \begin{subfigure}{0.3\linewidth}
    \includegraphics[trim=0 0 0 0, clip, width=\linewidth]{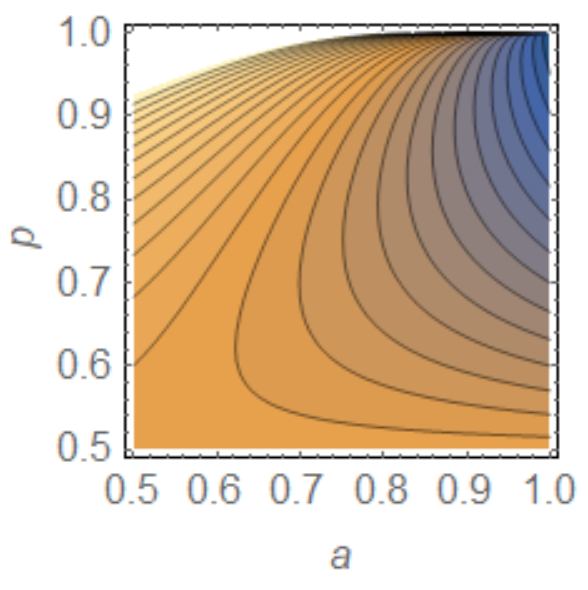}
    \vspace{-1.8em}
    \caption{$\alpha$-type}
    \label{fig:alpha equipotential}
    \end{subfigure}
    \hfill
    \begin{subfigure}{0.3\linewidth}
    \includegraphics[trim=0 0 58 0, clip,width=\linewidth]{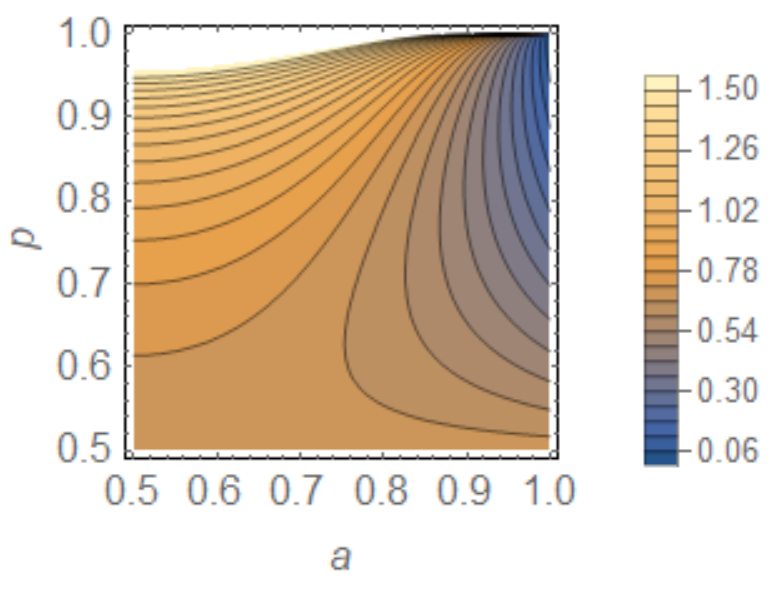}
        \vspace{-1.5em}
    \caption{$\beta$-type}
    \label{fig:beta equipotential}
    \end{subfigure}
    \hfill
    \begin{subfigure}{0.3\linewidth}
    \vspace{0.3em}
    \includegraphics[width=\linewidth]{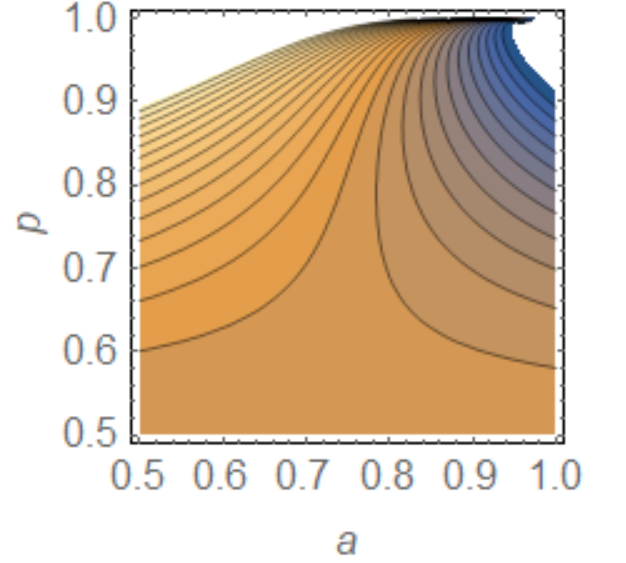}
        \vspace{-1.em}
    \caption{$\gamma$-type}
    \label{fig:gamma equipotential}
    \end{subfigure}
    \vspace{-0.5em}
    \caption{Visualized landscapes of the three different generalization losses we calculated in Equation (\ref{eq: alpha generalization loss 2class}), (\ref{eq: iid corruption generalization error}) and (\ref{eq: raw generalization error}). Regions of lower generalization loss are colored \textit{dark blue}. There are a few interesting points to note: (1) whenever $a\neq 1$, a divergence in the test loss exists when $p=1$; this is simply because $\log 0$ diverges in the loss; therefore, the foremost benefit of label smoothing (however minor) is that it prevents the divergence of the test loss; (2) for every fixed $a$, a unique optimal $p=p^*$ where the test loss is minimized exists (take a vertical slice of any of the above figure gives a unique point where the plot is darkest). For where these optimal points lie, see Figure~\ref{fig:2-class solution}.\vspace{-1mm}} %
    \label{fig: landscapes}
\end{figure*}

\textbf{True generalization loss}. Notice that the generalization loss defined above is the measured loss on the same data points as the training points, which disagrees with the standard definition of the test loss, which should be defined as the expected loss over the complete data distribution $p(x,y)$. However, when the number of training data points is large enough, the test error is well approximated by the test error measured on the same training set with correct labels. This can be justified by the standard VC-dimension related bound, where it is easy to show that the true generalization error of our model $h$, $R(h)$, can be bounded as $R(h)\leq R_{empirical}(h) + C(|H|, N)$, where $|H|$ is a constant measuring model complexity and $N$ the number of data points, and $C$ is a decreasing function of $N$. In practice and in this paper, the number of data points is larger than or equal to $10^4$, which makes $C$ very small, and the empirical test loss we defined is likely to offer a good estimate of the true testing loss. Since this is not the focus of this work, we do not go deeper into this issue. Discussions about the classical VC bound can be found in~\cite{shalev2014understanding}, and more recent and detailed bounds that apply to neural networks are studied in~\cite{arora2018stronger}.

\vspace{-1mm}
\section{Optimal Label Smoothing}
\vspace{-1mm}

Given this setup, we can now obtain the generalization loss in the setting outlined in the previous section. The calculation of the empirical generalization loss depends on our assumptions about the distribution of the test set; we study three kinds of assumptions about the testing set in each of the following 3 subsections. In Figure~\ref{fig: landscapes}, we show that three different assumptions regarding the test set results in three different types of generalization loss landscape in the $a-p$ plane. These landscapes feature different optimal solutions of $p$ that minimizes the generalization loss for fixed $a$ (see Figure~\ref{fig:2-class solution}). Moreover, in our framework, we show that, for every value of label smoothing $p$ and clean rate $a$, the empirical generalization loss can be calculated which is very useful in estimating the true generalization loss rather accurately.

\subsection{$\alpha$-Type Theory: A Clean Testing Set}

\textit{Assuming that the testing set is a non-corrupted set, with the true targets being one-hot.} This is particularly simple. The generalization loss is given by 
\begin{equation}\label{eq: alpha generalization loss 2class}
    -a\log(p) - (1-a)\log(1-p),
\end{equation}
which is minimized by $p^*_\alpha=a$, where the generalization loss is equal to $H(a)$, the entropy of the $a$-binomial distribution. We call this solution $p^*_\alpha$, see Figure~\ref{fig:2-class solution}. The $\alpha-$type noise agrees with what is commonly assumed in the literature of robust learning against noisy labels; where the training set is partially corrupted while the testing set is assumed to be clean~\citep{han2018co,patrini2017making,ziyin2020learning}.

\subsection{$\beta$-Type Theory: A Corrupted Testing Set}

\textit{Assuming that the testing set is corrupted by the same level of i.i.d. corruption in the training set.} The generalization error of this solution is 
\begin{equation}\label{eq: iid corruption generalization error}
    \ell(p,a) =  - a^2 \log p -2a(1-a)\log(1-p) - (1-a)^2 \log(p),
\end{equation}
and the solution $p^*$ becomes, interestingly, a non-linear function of $a$. An analytical solution exists, that is, $p^*_\beta=2a^2 -2a + 1$. Also note that this optimal rule results in a smaller $p$ than the optimal rule in the previous section: $p^*_\beta \leq p^*_\alpha$. In other words, when the testing set is also corrupted, \textit{we need stronger smoothing}. The landscape of the generalization loss in the above equation is plotted in Figure~\ref{fig: landscapes}, and $p^*_\beta$ is plotted in Figure~\ref{fig:2-class solution}.
The $\beta$-type assumption is a more realistic setting for a standard dataset than the $\alpha-$type, because, for commonly used datasets, the testing set and the training set are collected and labeled using the same procedure. Therefore, the dataset annotators are likely to exhibit the same rate of errors in both training and testing sets which implies that the training and the testing sets are likely to have the same amount of label noise.

\subsection{$\gamma$-Type Theory: A Bayesian Approach}

\begin{figure}
    \centering
    \includegraphics[width=0.4\linewidth]{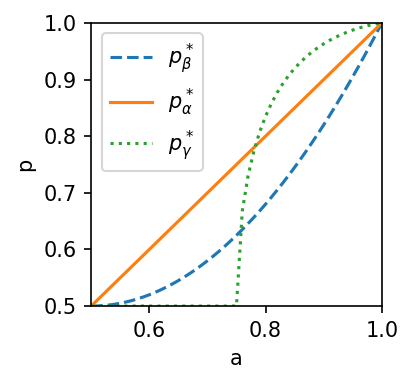}
    \vspace{-1em}
    \caption{{Optimal label smoothing $p^*$ for each clean rate $a$ in a 2-class setting, solved by minimizing $p$ over the generalization losses in Equation (\ref{eq: alpha generalization loss 2class}), (\ref{eq: iid corruption generalization error}) and (\ref{eq: raw generalization error}). On the one hand, the shared feature is that, when the corruption rate is small, smaller strength of label smoothing is needed; when the corruption rate is large, large label smoothing strength is needed. On the other hand, we see that three different kinds of assumptions lead to three distinctive optimal label smoothing rates. This also suggests that, before applying label smoothing, it is important for the practitioners to check whether the distribution for the test set to agree with the training set, and check which assumption applies more readily to their task at hand.}} %
    \vspace{-0em}
    \label{fig:2-class solution}
\end{figure}
We might also think of the smoothed targets $p$ as an \textit{prior assumption} about the conditional distribution of the target, i.e. we make no assumption about whether the labels are corrupted, but only an \textit{a priori} assumption about how its labels are distributed given an input. Additionally, we assume that the testing set and the training set are sampled from the same distribution. If this is the case, the testing loss should also be calculated with respect to the \textit{a priori} target $p$. The generalization error of this solution is then
\begin{equation}\label{eq: raw generalization error}
    \ell(p,a) = aH(p) - (1-a)\left[p\log(1-p) + (1-p)\log p \right]
\end{equation}
where we have taken expectations over the noise. We observe a bias-variance trade-off, where the first term $H(p)$ denotes the variance in the original labels, while the second term is the bias introduced due to noise. As $a\to 1$, the noise disappears and we achieve perfect generalization where the training loss is the same as generalization loss.

However, an analytical solution does not exist in this case. We thus resort to numerical calculation to solve this equation. See Figure~\ref{fig:2-class solution} for the solution. While the testing loss is minimized for any $a>0.75$, one interesting point is that the theory predicts uniformly $p^*$ in the range $0.5<a<0.75$. The $\gamma$ type assumption applies more often in language and sentiment analysis tasks. For example, upon hearing the sentence ``this movie is remarkably fine!'', the annotator is asked to label the sentiment of the spoken utterance; to some people, this sentence might appear positive; while to others, this sentence might appear sarcastic and, thus, negative, while depends highly on prior beliefs. There exists some inherent degree of uncertainty in the labels collected for these datasets which makes it important to study them while accounting for label noise\footnote{More detailed and related discussion can be found in~\cite{liang2018multimodal}.}.

\vspace{-1mm}
\section{Multiclass and Non-Uniform Label Smoothing}\label{sec: non-uniform transition matrix}
\vspace{-1mm}

In this section, we generalize our result to a multiclass setting. We consider the case when the corruption on the labels are not uniform, and our transition matrix $T\in \mathbb{R}^{M\times M}$ defines class-conditional noise \citep{denis2006efficient,natarajan2013learning,scott2013classification}. For example, $T_{ij}$ defines the probability of an clean label $j$ transitions to another label $i$ via noise. In expectation, a one-hot label $\tilde{y}_i$ corrupted by $T$ is equal to $\mathbb{E}[\tilde{y}_i] = Ty_i$. In fact, the results are not qualitatively different from the $2-$class case.

\subsection{$\alpha$-type, and forward-matrix correction}\label{sec: alpha type theory and forward matrix}

\begin{figure}
    \centering
    \includegraphics[width=0.4\linewidth]{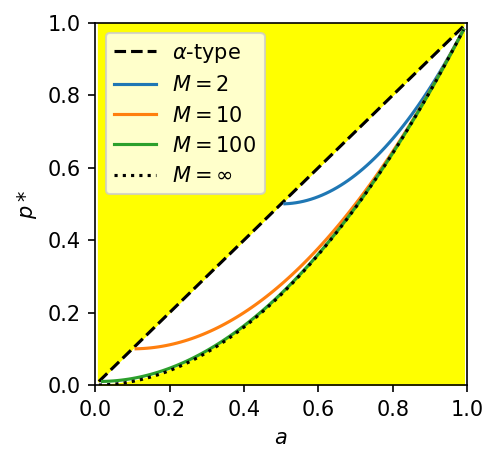}
    \vspace{-1.0em}
    \caption{{Optimal label smoothing rate $p^*$ for each clean rate $a$ for $\alpha$ and $\beta$-type assumptions in the multi-class setting with uniform noise. Two interesting points to note: (1) the $\alpha$-type theory suggests the optimal label smoothing rate regardless of the number of classes $M$; (2) the $\beta$-type solution converges to a limiting function as $M$ increases; and discernible difference with the limiting function disappears when $M>100$.}} %
    \label{fig:multiclass solution}
\end{figure}
The generalization loss for the first class can be written as
\begin{equation}
    \ell_\alpha^1(S, T) = -\sum_{j=0}^M T_{1j} \log(S_{1j}),    
\end{equation}
where we treat the network output $f$ as a vector, and $\log$ as an element-wise function to make the notation concise. The solution is simple, and can be generalized to any class $i$: $S^*_{ij} = T_{ij}$. When the transition matrix is uniform, i.e. a correct label remains correct with probability $a$, and corrupts to any other specific label with probability $\frac{1-a}{M-1}$, the generalization loss can be written as
\begin{equation}
\label{eq: alpha type multiclass equation}
    \ell_\alpha^1(p, a) =- a\log p - (1-a)\log \left(\frac{1-p}{M-1}\right).
\end{equation}
We can obtain the solution $p^*_\alpha=a$. The optimal $p^*$ does not depend on the number of classes $M$. Compared to the binary classification case, we see that the effect of multiclass classification shifts the generalization loss by a constant factor $(1-a)\log(M-1)\geq 0$, signalizing the fact that it is harder to defend against overfitting as the number of classes grows.

\textit{Relation to Forward-Matrix Correction} \citep{patrini2017making}. In fact, the above analysis shows that the optimal label smoothing rule obtained above is equivalent to a label corruption method called \textit{Forward-Matrix Correction} (F-matrix) \citep{patrini2017making} (which is in fact the state-of-the-art method for combating label corruption problems when $T$ is known) despite the fact that both methods are motivated and implemented very differently. This equivalence was also noticed by \cite{lukasik2020does} to argue that label smoothing might defend against label corruption. The F-matrix method proposes to correct the output of the model $f(x)$ by the noise transition matrix $f(x)\to Tf(x)$, while leaving the target label one-hot. While the solution to this method is $f^*(x_i) = \tilde(y_i)$, the corrected output becomes $Tf^*(x_i) = T\tilde{y}_i =S^*\tilde{y}_i$, and computing the generalization loss using this corrected output gives exactly the same value as the $\alpha$-type optimal label smoothing. However, the asymptotic solution has the same generalization loss. This equivalence in the solution suggests that label smoothing might be fundamentally invoking the same mechanism for improving generalization as the forward-matrix method. %

\subsection{$\beta-$type theory}

The generalization loss is (summed over all the classes)
{\fontsize{8}{12}\selectfont
\begin{equation}
    \ell_T = \sum_{i=1}^M \left[- T_{ii}^2 \log(S_{ii}) -  \sum_{j\neq i}^M T_{j,i}^2 \log(S_{jj})  - 2  \sum_{k\neq j}^M \sum_{j}^M T_{ki}T_{ji} \log(S_{kj}) \right].    
\end{equation}
}While this looks like a complicated system of equations to solve, we can again solve it by taking the derivative and setting it to $0$. Starting with the diagonal terms, we obtain that, for specific class $i$, $S^*_{ii}\sim T_{ii}^2 + \frac{1}{M-1}\Bar{T}_{/i}^2$, where we defined the \textit{off-diagonal average} $\Bar{T}_{/i}^2:= \frac{1}{M-1}\sum_{j\neq i }^M T_{ij}^2$ for both conciseness and insight. The probability that both the training set and testing set are corrupted to the same wrong class $j$ emits the correction term $\Bar{T}_{/i}^2$.

\begin{figure}
    \centering
    \includegraphics[width=0.4\linewidth]{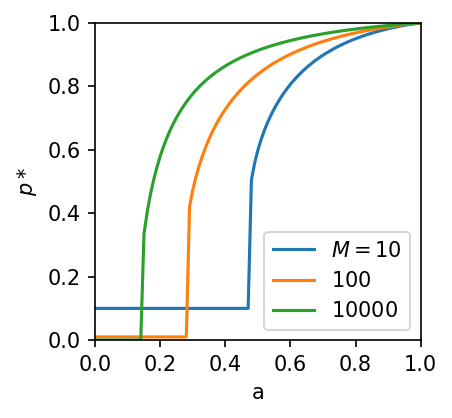}
    \vspace{-1em}
    \caption{{Solution for $\gamma$-type assumptions in multi-class and uniform noise setting. As is the case for $\alpha$ and $\beta$-type, qualitatively similar trend is observed for different $M$.}} %
    \label{fig:gamma multiclass solution}
\end{figure}
The solution to the off-diagonal terms are easier to obtain: $S^*_{ij}\sim 2 \sum_{k=1}^M T_{ik}T_{jk} $, and combining these results, we have that for any $i,j\in{1,...,M}$, the optimal solution is given by
\begin{align}
    S_{ij}^* = \frac{1}{2}\sum_{k=1}^M T_{ik} T_{jk},\quad\quad \textit{ or equiv., }  S^*= \frac{1}{2}T T^t.
\end{align}
It is interesting to note the difference between this solution and the $\alpha$-type solution: while the $\alpha$-type solution is a general doubly stochastic matrix, the $\beta$-type solution is a symmetric matrix. To our knowledge, the $\beta$-type solution has never explored in the literature (neither in the literature of label smoothing nor of the label corruption).%

When the noise is uniform, i.e., when the transition matrix takes the form $T_{ii}=a$, $T_{ij}=\frac{a}{M-1}$, the \underline{$\beta$-type} generalization loss takes a neat form:
{\fontsize{9.5}{12}\selectfont
\begin{align}
    \ell_\beta^{1,M}&(p,a) %
    = - a^2 \log(p) - 2a(1-a)\log\left(\frac{1-p}{M-1}\right) - \frac{(1-a)^2}{M-1}\log\left(p\right) - \frac{(1-a)^2(M-2)}{M-1} \log\left(\frac{1- p}{M-1}\right)
\end{align}
}Applying the general result, one obtains $p^*_\beta=\frac{1 - 2 a + a^2 M}{M-1}$. We plot this solution for different $M$ in Figure~\ref{fig:multiclass solution}. We note that there are two things interesting about this solution: (1) when corruption rate is small ($a\geq 0.8$), the solutions at different $M$ well approximate each other, and (2) as $M\to \infty$, the limit $\lim_{M\to \infty} p^*_M = a^2$ exists.

\subsection{$\gamma-$type theory}

Since no analytical solution exists for $\gamma$-type theory; we only numerically study the case when the noise is uniform. In this case, the generalization loss is given by the following:
{\fontsize{8}{12}\selectfont
\begin{align}
    \ell_\gamma^{1,M}(p,a)& = a[H(p)+(1-p)\log(M-1)] - (1-a)\bigg[ p \log\left( \frac{1-p}{M-1}\right) + \frac{(1-p)(M-2)}{M-1} \log\left(\frac{1-p}{M-1} \right) +\frac{1-p}{M-1}\log(p) \bigg],
\end{align}
}and so the solution is also dependent on $M$. While an analytical solution does not exist, one can hope to identify something interesting as $M$ becomes large. In this case, the generalization loss becomes $\ell=\ell_{2-class}(a,p) - ap\log(M-1) + O(\frac{1}{M}) + const.$, and so the difference with the $2$-class case comes from the term $-ap\log(M-1)$, which encourages larger $p$ as $M-1$ increases. We plot the numerical solution in Figure~\ref{fig:gamma multiclass solution}. The characteristic feature of the $\gamma$-type theory is a sharp, first-order transition for some $a>\frac{1}{M}$, where the global minimum becomes not learning (by setting $p=\frac{1}{M}$).

\begin{figure}[t]
    \begin{subfigure}{0.225\linewidth}
    \includegraphics[trim=0 0 0 0, clip, width=\linewidth]{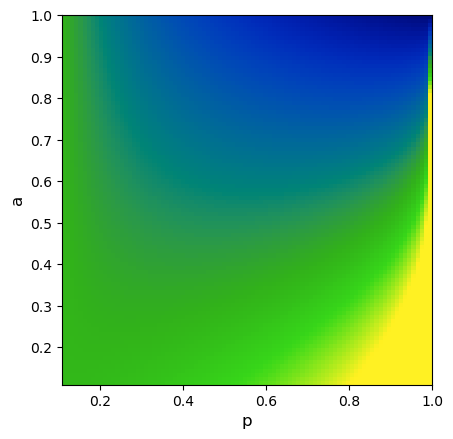}
    \caption{$\alpha$-type theory, MNIST}
    \end{subfigure}
    \hfill
    \begin{subfigure}{0.26\linewidth}
    \includegraphics[width=\linewidth]{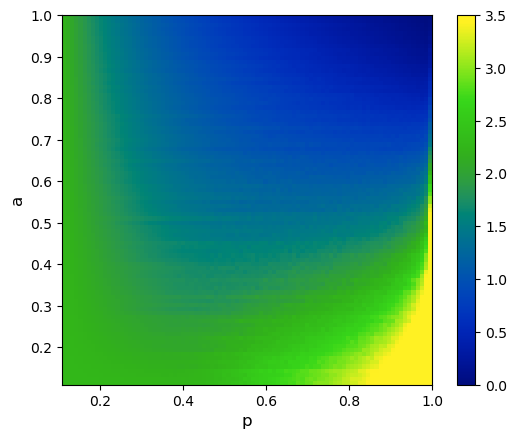}
    \caption{Empirical test loss, MNIST}
    \end{subfigure}
    \begin{subfigure}{0.225\linewidth}
    \includegraphics[trim=0 0 0 0, clip, width=\linewidth]{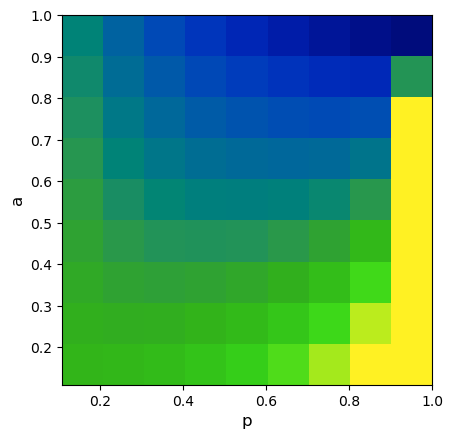}
    \caption{$\alpha$-type theory, CIFAR10}
    \end{subfigure}
    \hfill
    \begin{subfigure}{0.26\linewidth}
    \includegraphics[width=\linewidth]{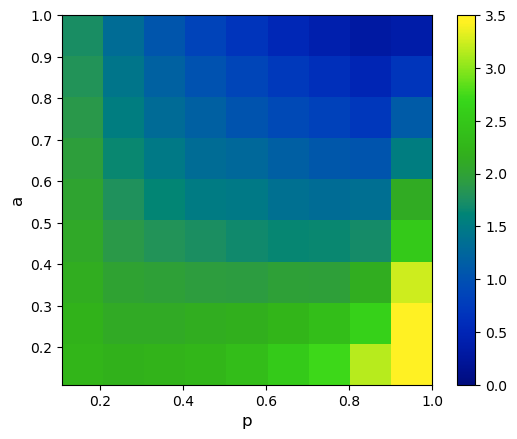}
    \caption{Empirical test loss, CIFAR10}
    \end{subfigure}
    \caption{Visualized landscape of the theoretical (a,c) and empirical (b,d) test loss $\alpha$-type conditions on the MNIST (a,b) and CIFAR10 (c,d) datasets with a granularity of $0.01$/$0.1$ for $a$ and $p$. The theoretical value on this $a-p$ grid is calculated from Equation (\ref{eq: alpha type multiclass equation}). The sheer visual similarity between the theoretical plot and the measured data justifies the correctness of our assumption about the testing set and, therefore, the $\alpha-$type theory that is established based on it.}
    \label{fig: alpha experiment}
\end{figure}

\subsection{``Mean-Field" Label Smoothing for General Corruption Matrix $T$}

The most general methods developed in the previous section, however, hinges on knowing the transition matrix $T$, which is also an often criticized drawback of the F-matrix method~\citep{ziyin2020a,ziyin2020learning}. In practice, there does not exist a widely-accepted method to accurately estimate the noise transition matrix\footnote{Of course, some methods do exist~\citep{patrini2017making, aaai19jiangchao}.}. In many such cases, it is much easier if the label corruption can be summarized by a simple scalar $a$, measuring the averaged clean rate, and the $1-a$ can be understood as an effective, or \textit{mean-field}, corruption rate. If we can achieve this, the problem becomes much simpler, and one additional advantage is that this also agrees with the practical case where the label smoothing is applied in its most basic form, i.e., a $p$-smoothing. 

In this case, the generalization loss and the solutions are, respectively,
\begin{equation}
   \ell_T^1 = \begin{cases} -\sum_{j=0}^M T_{1j} \log(S_{1j}), & \text{($\alpha$-type)}\\ 
  - T_{11}^2 \log(p) - \sum_{j\neq 1}^M T_{j1}^2 \log(p) \\ 
  \>\>\>\>\>\>- 2 \sum_{k\neq j}^M \sum_j^M T_{k1}T_{j1} \log(1-p) & \text{($\beta$-type)}
    \end{cases}
\end{equation}
\begin{equation}
    \to \begin{cases}p^*_\alpha=\frac{1}{M} \sum_{i=1}^M T_{ii}\\ 
    p^*_\beta= \frac{1}{2}\sum_{i=1}^M \sum_{j=0}^M T_{ij}^2 = \frac{1}{2M}\mathrm{Tr}[TT^t].
    \end{cases}
\end{equation}
Thus, $\bar{a}=\frac{1}{M} \sum_{i=1}^M T_{ii}$ can be defined as an \textit{effective} clean rate. These mean-field approximations are extremely useful when $T$ is poorly estimated, for example, due to insufficient sampling. For example, for the ImageNet dataset~\citep{imagenet_cvpr09}, $M=1000$, and estimating $T$ involves estimating $1,000,000$ variables, while estimating $\bar{a}$ requires estimating only $2$ variables. Given that ImageNet has only $1,000,000$ data points, there is no hope one can estimate such a high-dimensional matrix due to the curse of dimensionality~\citep{hastie2009elements}. On the other hand, computing the effective clean rate requires one more averaging over the $M$ classes, and is thus much more accurate than the estimated individual elements in the transition matrix. Many practical methods also require a knowing a single variable $a$ rather than the full transition matrix, such as the pruning rate in~\cite{han2018co} or the early stopping criterion in~\cite{han2018co} and~\cite{ziyin2020learning,ziyin2020a}. We note that there is not yet a simple yet effective method to estimate the effective clean rate, and this can be of great value to future research.

\begin{figure}
\begin{subfigure}{0.24\linewidth}
    \includegraphics[width=\linewidth]{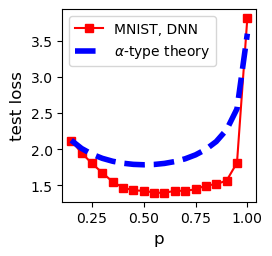}
    \caption{$a=0.50$}
\end{subfigure}
\begin{subfigure}{0.24\linewidth}
    \includegraphics[width=\linewidth]{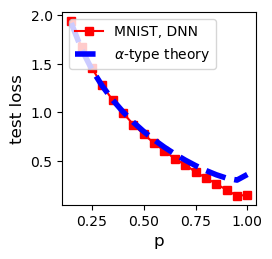}
    \caption{$a=0.95$}
\end{subfigure}
\begin{subfigure}{0.24\linewidth}
    \includegraphics[width=\linewidth]{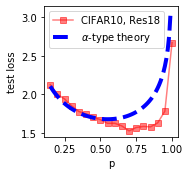}
    \caption{$a=0.50$}
\end{subfigure}
\begin{subfigure}{0.24\linewidth}
    \includegraphics[width=\linewidth]{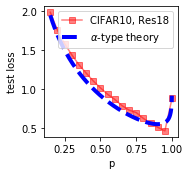}
    \caption{$a=0.95$}
\end{subfigure}
\caption{Plotted is the predicted and measured test loss $\ell(a,p)$. Agreement with $\alpha$-type theory is good both in trend and in absolute number.\vspace{-4mm}}
\label{agreement between theory and experiment}
\end{figure}

\vspace{-2mm}
\section{Experiments}
\vspace{-2mm}

In this section, we demonstrate the validity of our theories via a set of empirical experiments on datasets with varying levels of corruption. Unlike our theoretical calculations, in practical settings, the measured test loss is calculated on a held-out set of data points that do not overlap with the training set. Therefore, the agreement between our theory and the measured test loss will justify our assumption that the empirical generalization loss is a good estimate of the true generalization loss. The code for our experiments will be released after the blind review period. Please refer to the Appendix for implementation details and the set of hyperparamters used.

\subsection{$\alpha$-Type Noise}

To confirm the $\alpha$-type theory, we conduct experiments with a corrupted training set at a clean rate of $a$ and a clean hold-out test set. We manipulate the clean rate $a$ of the datasets by introducing \textit{symmetric noise} into the datasets, where corrupted training data have their labels randomly flipped to a different label. In particular, in a classification task with $M$ classes, any given data point $(x_i, y_i)$ had $y_i$ perturbed to $y'_i$, where $P(y'_i=y_i)=a$ and $P(y'_i=y)=\frac{1-a}{M-1}$ for $y\neq y_i$. We perform a grid search, training a neural network with varying smoothing parameters $p$ and clean rates $a$ in order to empirically determine the test loss given a particular clean rate and smoothing parameter. Our findings show that the empirical test loss closely matches the test loss predicted by the $\alpha$-type theory. See Figure~\ref{fig: alpha experiment}. We first compare visually and qualitatively. We see that for both MNIST and CIFAR-10 dataset, the predicted test loss agree very well with the measured test loss. The visual similarity suggests the agreement between theory and our experiment.

To allow for more precise comparison, we also fix $a$ take two slices from these plots and compare theory and experiments (see Figure~\ref{agreement between theory and experiment} for 2 commonly tested values of $a$). We observe high agreement in both the trend and in the absolute value of predicted vs measured test loss. We also observed similar high degrees of agreement for other values of $a$. The agreement between our theory and the measured test loss justifies our assumption that the empirical generalization loss is a good estimate of the true generalization loss.

\begin{figure}[b]

\begin{subfigure}{0.235\linewidth}
    \includegraphics[width=\linewidth]{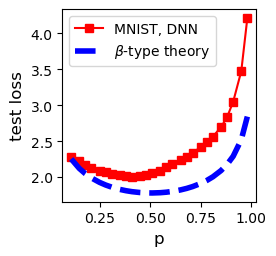}
    \caption{$a=0.50$}
\end{subfigure}
\begin{subfigure}{0.235\linewidth}
    \includegraphics[width=\linewidth]{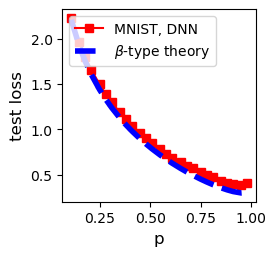}
    \caption{$a=0.95$}
\end{subfigure}
\centering
\begin{subfigure}{0.235\linewidth}
    \includegraphics[width=\linewidth]{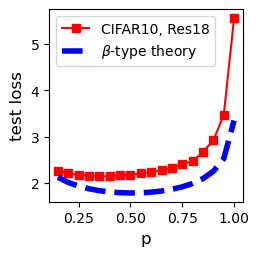}
    \caption{$a=0.50$}
\end{subfigure}
\begin{subfigure}{0.235\linewidth}
    \includegraphics[width=\linewidth]{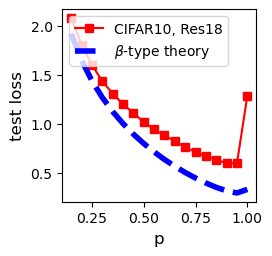}
    \caption{$a=0.95$}
\end{subfigure}
\caption{Plotted is the predicted and measured test loss $\ell(a,p)$ on MNIST and CIFAR10, showing good agreement with $\beta$-type theory.}
\label{beta cifar}
\end{figure}

\subsection{$\beta$-Type Noise}

Now we study the case that when $\beta-$type noise is present in our dataset. For a given clean rate of $a$, we corrupt both the training and the testing set with the same but independent uniform corruption, as is described in the theory section.
We confirm $\beta$-type theory with a set of experiments at specific clean rates, where the clean rates are applied to the held-out test set as well. We test the similarity of our derived $\beta$-type theory with empirical results with the specific clean rates of $0.5$ and $0.95$ on a set of different parameters to demonstrate the predictive power of our $\beta$-type theory. See Figure~\ref{beta cifar} for the experiment on MNIST and CIFAR10.  Our findings again confirm the close correspondence between our derived theory and the empirical test loss on real datasets. This suggests that our theory provides a very good first-order estimate of the generalization loss of the label smoothing method.

\vspace{-1mm}
\section{Discussion}
\vspace{-1mm}

In this section, we discuss some implications of this work and interesting open problems we discovered about label noise but do not yet understand. 

\subsection{Discrepancy between test loss and test accuracy}

\begin{figure}
    \centering
    \includegraphics[width=0.4\linewidth]{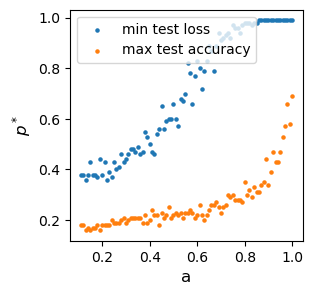}
    \vspace{-1em}
    \caption{Optimal $p^*$ for minimizing test loss and maximizing test accuracy given $a$ on MNIST. Note the wide discrepancy between the two patterns. In particular, a smaller smoothing parameter seems to provide a better test accuracy.}
    \vspace{-1em}
    \label{fig:discrepancy}
\end{figure}

Although our theories predict test loss and the optimal smoothing parameter well, we notice a surprising discrepancy between the minimizer of the test loss and the maximizer of the test accuracy in practice. See at Figure~\ref{fig:discrepancy}. While $p^*$ and $a$ are positively correlated for both curves, it is clear that there is actually a wide discrepancy between the optimal smoothing parameter for minimizing test loss and that for maximizing test accuracy. This surprising result runs counter to an implicit neural network assumption that optimizing for test loss should achieve near-optimal test accuracy, suggesting that, although we have captured the relationship in how label smoothing controls for test loss, there still remains a significant unexplored relationship between test accuracy and label smoothing that may further help label smoothing improve the performance of neural networks. This is an open problem that requires a solution in the future.

\subsection{Possibility of a new smoothing method}

In section~\ref{sec: alpha type theory and forward matrix}, we mentioned the possibility that the $\alpha$-type theory might share the same underlying mechanism with the forward-matrix method. Formally, the difference between label smoothing and forward-matrix is captured by Jensen's inequality: Let $h_i$ denote the pre-softmax values of the model prediction. Then the relationship between label smoothing and forward-matrix is (let $\sigma(\cdot)$ denote the commonly used soft-max function)
\begin{equation}
    \underbrace{\sum_{ij}^M S_{ij} \log(\sigma(h_j))}_{\text{label smoothing}} \leq \underbrace{\sum_{ij}^M \log(S_{ij}\sigma(h_j))}_{\text{forward-matrix}},
\end{equation}
which follows from Jensen's inequality. This suggests that we can also imagine a different version of smoothing by applying Jensen's inequality one more time:
\begin{equation}
    \underbrace{\sum_{ij}^M S_{ij} \log(\sigma(h_j))}_{\textrm{label smoothing}} \leq \underbrace{\sum_{ij}^M \log(S_{ij}\sigma(h_j))}_{\textrm{forward-matrix}} \leq \underbrace{\sum_{ij}^M \log(\sigma(S_{ij}h_j))}_{\textrm{logit smoothing}},
\end{equation}
This method of smoothing can be seen as \textit{logit smoothing}. Does logit smoothing work? It is beyond the scope of this work, and we leave it for future exploration.

\vspace{-1mm}
\section{Conclusion}
\vspace{-1mm}

In this work, we proposed the first theoretical framework to quantitatively explain the label smoothing technique in deep learning. Our theory points towards its effectiveness in controlling for generalization loss and also allows us to derive an \textit{optimal label smoothing} point for best performance. We verified these claims through extensive controlled experiments across several datasets and models under several realistic label noise settings.
There are also many other questions that remain open. For example, can label smoothing help when there is no corruption or noise in the labels? If so, is there a theoretical framework in which we can understand it? We hope this work will stimulate more interest in studying the label smoothing problems from theoretical and empirical perspectives.

\bibliographystyle{plain}
\bibliography{ref}

\clearpage

\appendix

\onecolumn

\clearpage
\section{Additional Experiments}

\subsection{Detailed Plots of the MNIST}
\begin{figure}[b!]
\centering
    \begin{subfigure}{0.45\linewidth}
    \includegraphics[trim=0 0 0 0, clip,width=1\linewidth]{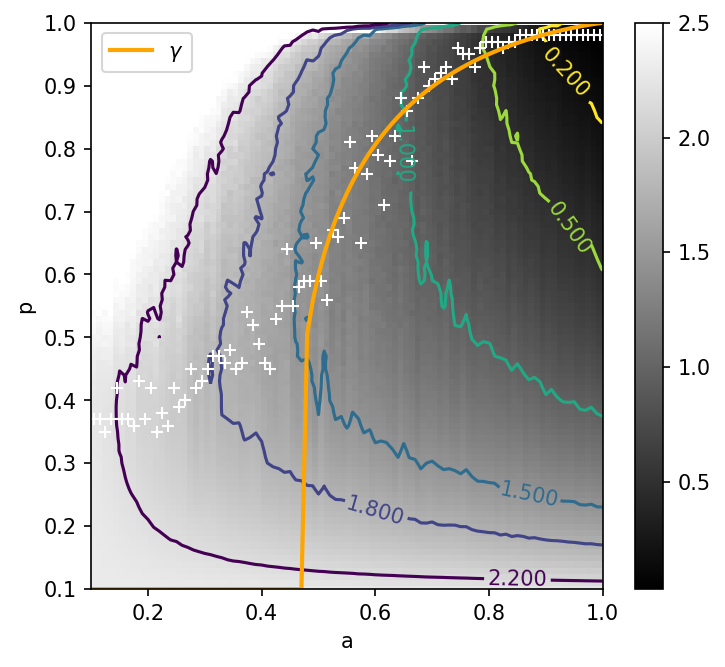}
        \vspace{-2.em}
    \caption{ Measured Testing Loss.}
    \end{subfigure}
    \begin{subfigure}{0.45\linewidth}
    \includegraphics[trim=0 0 0 0, clip, width=\linewidth]{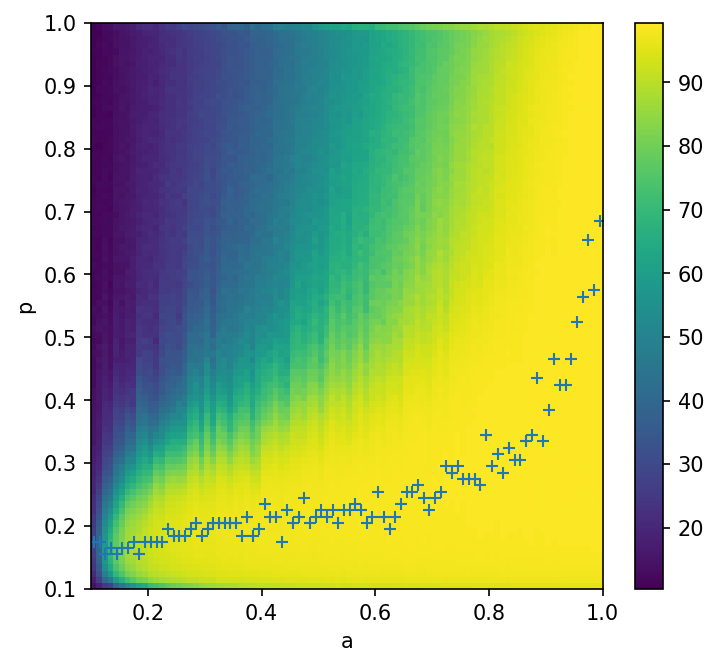}
    \vspace{-2.em}
    \caption{Final Performance.}
    \end{subfigure}
    
    \begin{subfigure}{0.4\linewidth}
    \includegraphics[trim=0 0 0 0, clip, width=\linewidth]{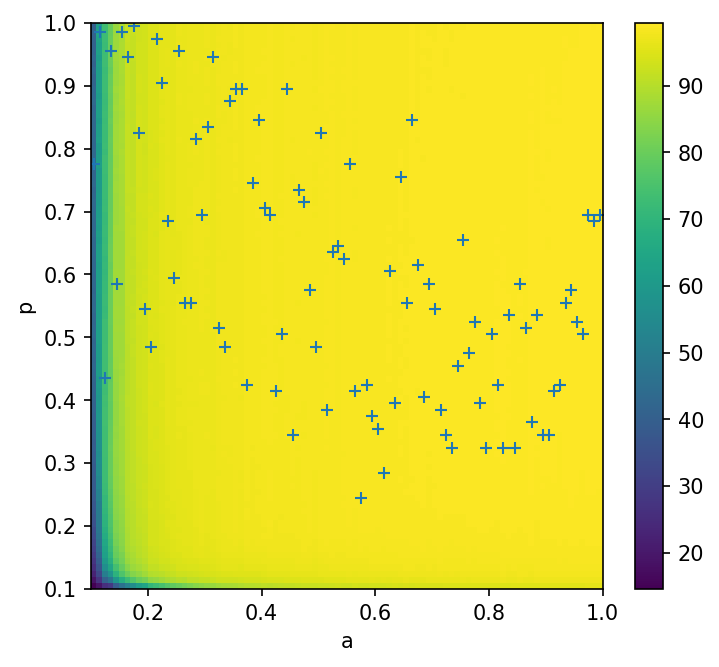}
    \vspace{-2.em}
    \caption{Performance At Optimal Early Stopping.}
    \end{subfigure}

    \vspace{-0.5em}
    \caption{\small{How label smoothing affects generalization accuracy. (a) Measured testing loss and the experimental minima (in white $+$ mark); $\gamma$-type theory predicts this kind of minima the best. (b) The test accuracy and the maximizers of the test accuracy (in blue $+$ mark); the fact that the minimizers of test loss does not agree with the maximizers of the test accuracy is very interesting, and suggests that a different theory is required to understand the trend in the the generalization accuracy. (c) One more interesting observation is that, at the early stopping point, the maximizer of the accuracy has a large dispersion, suggesting that it is not a function of the label smoothing value $p$ we used; this might imply that label smoothing is not needed if one performs early stopping. }}
    \label{app fig: gridsearch}
\end{figure}

Full plots for more detailed comparison on the experiments we have done on MNIST are given in this section. See Figure~\ref{app fig: gridsearch}. In addition to what we discussed in the main text, we also plot the generalization accuracy at the early stopping point, which also looks interesting; we see that at the early stopping point, the maximizer of the accuracy has a large dispersion, suggesting that it is not a function of the label smoothing value $p$ we used; this might imply that label smoothing is not needed if one performs early stopping. We plan to study this in a future work.%

\clearpage

\section{Non-Monotonicity in $p$: complicated tradeoffs}
\begin{wrapfigure}{rb}{0.4\linewidth}
\centering
    \begin{subfigure}{\linewidth}
    \centering
    \includegraphics[trim=0 0 0 0, clip,width=1\linewidth]{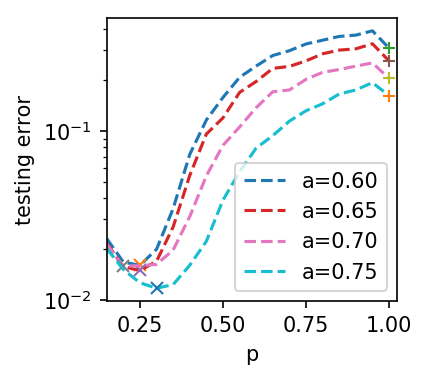}
        \vspace{-1.2em}
    \end{subfigure}
    \vspace{-0.5em}
    \caption{\small{ For $a\leq 0.85$, two minima exist in the error-$p$ curve. The global minimum in test error is achieved at small $p$, while no-smoothing constitutes a local minimum on a rather wide plateau.}}
    \label{fig: non-monotonicity}
\end{wrapfigure}

\begin{figure}[t!]
    \begin{subfigure}{0.35\linewidth}
    \includegraphics[trim=0 0 0 0, clip, width=\linewidth]{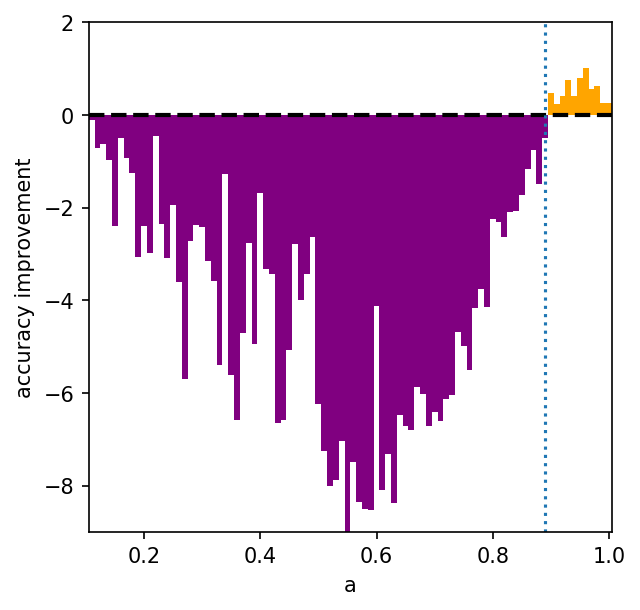}
    \vspace{-2.em}
    \caption{$p=0.99$.}
    \end{subfigure}
    \hfill
    \begin{subfigure}{0.31\linewidth}
    \includegraphics[trim=35 0 0 0, clip,width=\linewidth]{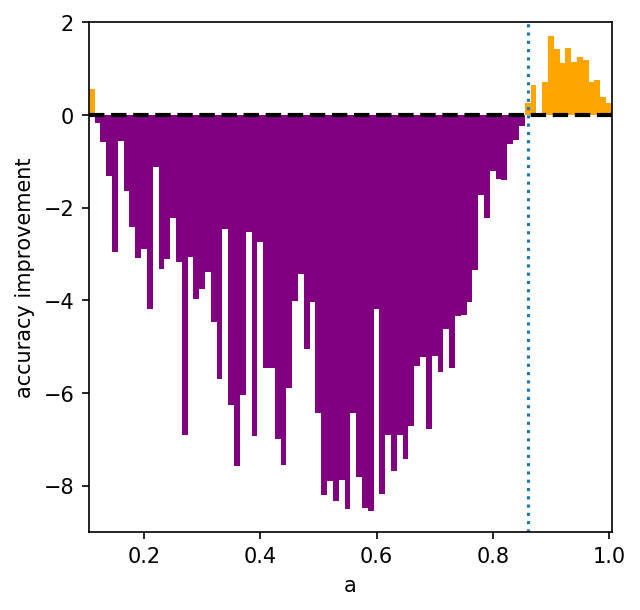}
        \vspace{-2.em}
    \caption{$p=0.95$.}
    \end{subfigure}
    \begin{subfigure}{0.31\linewidth}
    \includegraphics[trim=35 0 0 0, clip,width=1\linewidth]{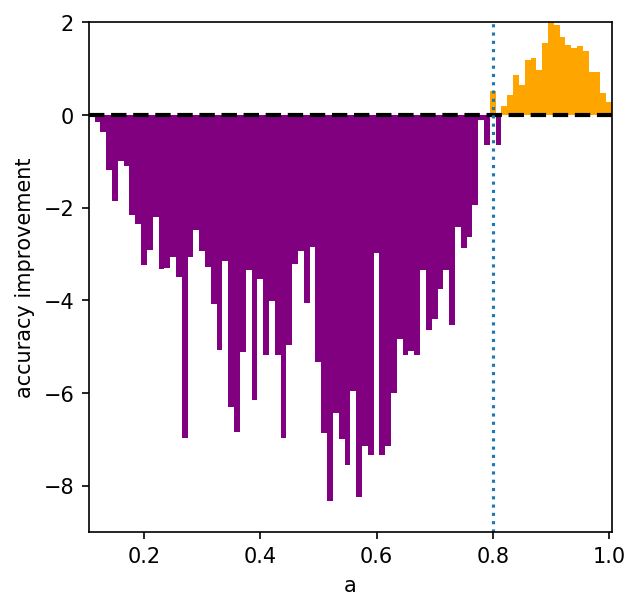}
        \vspace{-2.em}
    \caption{$p=0.90$.}
    \end{subfigure}
    
    \vspace{-0.5em}
    \caption{\small{How weak label smoothing ($p\sim 0.95$) affect generalization accuracy on MNIST. In fact, applying label smoothing might deteriorate generalization. See the text for detailed discussion.}}
    \label{app fig: mnist improvement}
\end{figure}

Experimentally, we demonstrate another phenomenon useful to the application of label smoothing. See Figure~\ref{fig: non-monotonicity}. Commonly $p$ is set to $0.95$ or $0.9$ in applications without much tuning, possibly due to the poor understanding of label smoothing previously. However, this experiment shows that using weak label smoothing $p~0.95,\ 0.9$ is only beneficial to learning at for a very small range of $a$ roughly from $a=0.9 \to 1.0$. For larger noises, the effect of label smoothing is detrimental, causing upto to $8\%$ accuracy drop compared to not applying to label smoothing at all; see Figure~\ref{app fig: mnist improvement}. 
This suggests that applying label smoothing is not as straightforward as we might have expected. In Figure~\ref{fig: non-monotonicity}.b, we plot $4$ examples of testing error vs. $p$ curve, and two minima are clearly observed in the curve, one at very small value of $p$, while the other at $p=1$. To our best knowledge, this work is the first to notice such effect. Again, we see that simply using \textit{weak} label smoothing does not result in straight forward improvement. %

Speaking in the ``confidence" language\footnote{Since label smoothing is said to prevent over-confident predictions.}, the lesson is that being overconfident ($p=1$) may in fact prevent overfitting, since being overconfident in some of data points will prevent one from formulating a theory to overfit to those points that do not make sense. We think understanding this effect theoretically will be very beneficial for our understanding of neural networks, and deserves a close study in future works. Experimentally, this suggests that the practitioners might want to apply label smoothing more carefully.

\section{Model Training Details}

Experiments were run on MNIST and CIFAR10. The CIFAR10 experiments were run on ResNet18 (1,000,000 parameters, \citep{he2016deep}), and the MNIST experiments was run on a custom CNN with $2$ convolutional layers and $2$ linear layers. Experiments were run for $100$ epochs each with a batch size of $128$ and a learning rate of $0.01$. SGD optimizers were used with a momentum of $0.9$ and no weight decay. For CIFAR10 only, the learning rate was halved after $50$ and after $75$ epochs. The goal of these hyperparameters were to reach as close to convergence as possible on the training set by the end of training.

\end{document}

%% file: tables/literature_survey.tex
\begin{table*}[t]
\caption{Survey of literature label smoothing results according to \cite{muller2019does}.}\label{tab: survey}
\begin{center}
\vspace{-4.5mm}
\begin{tabular}{ccccc}

\hline
Data set & Architecture & Metric & Value w/o LS & Value w/ LS\\
\hline
ImageNet & Inception-V2~\citep{szegedy2016rethinking} & Top-1 Error & 23.1 & \textbf{22.8}\\
ImageNet & Inception-V2~\citep{szegedy2016rethinking} & Top-5 Error & 6.3 & \textbf{6.1}\\
EN-DE & Transformer~\citep{vaswani2017attention} & BLEU & 25.3 & \textbf{25.8}\\
WSJ & BiLSTM+Att.~\citep{chorowski2016towards} & WER & 8.9 & \textbf{6.7}\\
\hline
\end{tabular}
\end{center}
\end{table*}